\documentclass[conference, 10pt]{IEEEtran}
\IEEEoverridecommandlockouts

\usepackage{cite}
\usepackage{amsmath,amssymb,amsfonts}
\usepackage{algorithmic}
\usepackage{graphicx}
\usepackage{tabularx}
\usepackage{url}
\usepackage{times}
\usepackage{float}
\usepackage{eso-pic}

\usepackage{booktabs}
\usepackage{textcomp}
\usepackage{xcolor}
\def\BibTeX{{\rm B\kern-.05em{\sc i\kern-.025em b}\kern-.08em
    T\kern-.1667em\lower.7ex\hbox{E}\kern-.125emX}}
\begin{document}

\title{Behavioral Fingerprinting and Navigation Prediction in Web Browsing}

\author{\IEEEauthorblockN{  Ralph Elsaghbini}
\IEEEauthorblockA{\textit{Computer sciences and networks department (INFRES)} \\
\textit{Institut Polytechnique de Paris, Telecom Paris, }\\
Palaiseau, France \\
ralph.elsaghbini@telecom-paris.fr}
\and
\IEEEauthorblockN{ Omran Berjawi}
\IEEEauthorblockA{\textit{Computer sciences and networks department (INFRES)} \\
\textit{Institut Polytechnique de Paris, Telecom Paris, }\\
Palaiseau, France \\
omran.berjawi@telecom-paris.fr}
\and
\IEEEauthorblockN{ Walid Fahs}
\IEEEauthorblockA{\textit{Faculty of Engineering} \\
\textit{Islamic University of Lebanon, Faculty of Engineering}\\
Wardanieh, Lebanon \\
walid.fahs@iul.edu.lb }
\and
\IEEEauthorblockN{ Rida Khatoun}
\IEEEauthorblockA{\textit{Computer sciences and networks department (INFRES)} \\
\textit{Institut Polytechnique de Paris, Telecom Paris, }\\
Palaiseau, France \\
rida.khatoun@telecom-paris.fr}
}

%{\footnotesize \textsuperscript{*}Note: Sub-titles are not captured for https://ieeexplore.ieee.org  and should not be used}

%\IEEEoverridecommandlockouts
      
%\IEEEoverridecommandlockouts\IEEEpubid{\makebox[\columnwidth]{979-8-3195-1282-6/26/\$31.00~\copyright~2026 IEEE} \hspace{\columnsep}\makebox[\columnwidth]{ }}

\AddToShipoutPictureBG*{%
  \AtPageLowerLeft{%
    \raisebox{0.6cm}{%
      \makebox[\paperwidth]{%
        \begin{minipage}{\textwidth}\centering\scriptsize
          \copyright~2026 IEEE. Personal use of this material is permitted.
          Permission from IEEE must be obtained for all other uses, in any
          current or future media, including reprinting/republishing this
          material for advertising or promotional purposes, creating new
          collective works, for resale or redistribution to servers or lists,
          or reuse of any copyrighted component of this work in other works.
        \end{minipage}%
      }%
    }%
  }%
}

\maketitle

\begin{abstract}
Web browsing often appears ephemeral: users visit a few websites, complete a task, and move on. However, even short fragments of browsing activity can contain rich and structured behavioral signals. In this work, we conduct a comparative empirical study of two complementary behavioral inference tasks: session-level user identification and next-domain prediction. Both tasks are derived from the same cleaned event stream and evaluated on large-scale anonymous browsing traces, with sessionization and splitting adapted to the
temporal requirements of each task. For user identification, we evaluate classical and neural models operating on session-level behavioral and domain features. For next-domain prediction, we combine graph-based modeling with Large Language Models (LLMs). Experimental results show that short browsing sessions are highly identifiable, while future navigation actions are highly predictable from long-term interaction structure combined with recent behavioral context. Furthermore, LLM-derived semantic features yield only marginal gains over purely structural and sequential models, indicating that repeated interaction patterns remain the
dominant predictive signal in the evaluated web-browsing setup. These findings highlight the extent to which interaction history substantially contributes to both user identifiability and navigation predictability in browsing traces.

\end{abstract}

\begin{IEEEkeywords}
Web browsing behavior, user identification, behavioral predictability, and privacy risk.
\end{IEEEkeywords}

\section{Introduction}
Web browsing is one of the most ubiquitous human behaviors in digital contexts. Users constantly engage with different sources, like news websites, search engines, social networks, and other web services, using short and disconnected browsing sessions. Despite their apparent ephemeral nature, such human behaviors emerge from consistent processes at various temporal scales. Previous research has proven the presence of significant regularities in human browsing behavior, which are formed due to human habits and repeated visits to certain websites~\cite{Kulshrestha2021ICWSM,SchnauberStockmann2023Predictability}. This suggests that even seemingly noisy browsing traces may encode structured behavioral signals.

Two main research perspectives have been developed to explore this structure: First, there is behavioral predictability that concerns how accurately future behaviors can be predicted from previous ones. The common approach for this line of work is sequential analysis of navigational patterns by predicting the next visited domain from the previous one \cite{Kulshrestha2021ICWSM, Paulino2024WebTraceSense}. Second, there is behavioral identifiability that considers whether particular individuals could be uniquely identified from their navigation behavior and domain preferences. This perspective has also become known as behavioral fingerprinting or re-identification \cite{Oliveira2025BehavioralFingerprinting, Song2025WebsiteFingerprinting}.

Although both identifiability and predictability originate from user domain interaction patterns, they are typically studied independently across prior work. Such an approach to behavioral analysis causes certain limitations because short-term temporal dynamics are commonly analyzed for predictability and long-term dynamics for identifiability. In addition, the findings within both perspectives are highly sensitive to methodological choices, including the process of sessionization, temporal splitting and representationlearning~\cite{Kulshrestha2021ICWSM,Paulino2024WebTraceSense}. Thus, it becomes hard to identify how much of the observed structure is related to the actual behavioral regularities.

In this paper, we study two complementary behavioral inference tasks derived from web browsing activity: session-level user identification and next-domain prediction. Our objective is to systematically compare how classical machine learning models, neural architectures, graph-based methods, and LLM-augmented approaches behave across these two related tasks. For the user identification task, each session is transformed into a fixed-dimensional representation composed of behavioral statistics, visited-domain information, and demographic-related features. We then evaluate multiple classification models, including linear, tree-based, neural, and representation-learning approaches, to measure how strongly user identity is encoded in short browsing activity. 

For the next-domain prediction task, we propose and evaluate the Structure-Dominant Hybrid Behavioral Model (SD-HBM), which combines graph representations and LLM-derived semantic features for  next-domain prediction. We evaluate all models on the same underlying event stream, using a common preprocessing pipeline with task-specific sessionization and splitting. For next-domain prediction the protocol is leakage-safe by construction, with chronological per-user splits and an interaction graph built exclusively from training sessions. This allows us to systematically quantify how different modeling families exploit the same behavioral signals and to assess the relative contribution of structural, sequential, and semantic information in browsing traces. The contributions of this work are as follows:

\begin{itemize}
    \item We conduct a comparative empirical study of two behavioral inference tasks derived from web browsing traces: session-level user identification and  next-domain prediction.  
     
    \item We evaluate classical, neural, graph-based, and LLM-augmented models under a consistent experimental pipeline.
    \item We analyze the role of structural interaction patterns in both identity inference and navigation prediction.
    
    \item We quantify the effect of LLM-based semantic augmentation across different integration strategies.
\end{itemize}

The remainder of this paper is organized as follows. Section~\ref{sec:related} reviews related work. Section~\ref{sec:framework} introduces our behavioral inference framework and describes the modeling approaches. Section~\ref{sec:experiments} presents the experimental setup. Section~\ref{sec:results} reports results. Section~\ref{sec:discussion} and Section~\ref{sec:limitations} discuss the results and the limitations. Section~\ref{sec:conclusion} concludes the paper.

\section{Related Work}
\label{sec:related}
Web browsing behavior has been researched within multiple communities, including web science, recommendation systems, user modeling, and privacy studies.

Web browsing behavior is highly regular over time: users repeatedly revisit a small set of websites, forming stable habits. Kulshrestha et al.~\cite{Kulshrestha2021ICWSM} demonstrate that browsing behavior is highly predictable but highly dependent on the methodology applied, including the way sessions are identified, pre-processing techniques, and the representation of features. Research in media psychology similarly reports stable browsing routines, with stronger habit associated with higher online predictability~\cite{SchnauberStockmann2023Predictability}.
 
Significant effort has been devoted to sequence and interaction modeling for next-action predictions. While earlier solutions based on Markov models and recurrent neural networks have been proposed, later research employed attention and Transformer-based architectures to enable the capturing of more long-range dependencies among interactions~\cite{Wang2021CSURSessionSurvey,Boka2024SeqRecSurvey,Pan2026SeqRecSurvey}. However, as shown empirically, there is no guarantee that the use of complicated architectures leads to an improvement in performance~\cite{LatifiMauroJannach2021InfSci,LatifiJannachFerraro2022InfSci}.

In addition to sequential models, some techniques consider higher-order structure in interaction datasets. Motif-based techniques capture repeated patterns of user activity~\cite{CuiEtAl2021MotifSIGIR}, while representation learning approaches leverage long-term user preference modeling using multi-session clickstream data~\cite{BlackEtAl2024TRACE}. This work underscores the notion that browsing behavior is characterized not only by immediate dependencies but by multi-scale structure.
 
In parallel to prediction-oriented efforts, other literature explores whether browsing behavior could lead to the identification of an individual. Existing literature highlights that a user’s browsing history and visited domains could serve as a fingerprint and make it possible to perform re-identification based on partial trace evidence~\cite{Oliveira2025BehavioralFingerprinting}. Privacy literature further highlights that user behaviors are a very sensitive type of attribute, which can be revealed even via aggregation or anonymization techniques. Analysis of privacy-preserving advertising mechanisms, including Google’s Topics API, shows that it is possible to launch successful re-identification attacks~\cite{BeuginMcDaniel2024TopicsAssessment}. Measurement results indicate that browsing behaviors can still leak user information even without direct identifiers in place~\cite{VernaEtAl2024CoNEXT}.

Our work performs controlled experiments to analyze how different modeling families, including classical machine learning models, graph-based methods, neural sequence models, and LLMs, capture behavioral regularities in web browsing traces.

\section{Behavioral Modeling Framework}
\label{sec:framework}
We model web browsing traces as structured behavioral signals that combine persistent user-specific patterns with transient contextual dynamics.
\subsection{Browsing Traces as Behavioral Signals}
We consider a set of browsing events $(u_i, d_i, t_i)$, where each event corresponds to a user $u_i$ visiting a domain $d_i$ at time $t_i$. These events form temporally ordered traces that reflect repeated interactions between users and web domains as a structured behavioral process:

\begin{itemize}
    \item \textbf{Persistent structure:} stable user-specific preferences that manifest as repeated visitation patterns over time.
    \item \textbf{Transient structure:} short-term contextual effects that influence immediate navigation decisions within sessions.
\end{itemize}

To capture these dynamics, we segment browsing traces into sessions using a time-gap rule. A session $S = (d_1, d_2, \dots, d_T)$ represents a contiguous episode of activity that reflects both habitual behavior and immediate intent.

\subsection{Two Complementary Inference Tasks}

We study two inference tasks that probe different aspects of the same underlying behavioral structure: one focuses on user identifiability, while the other focuses on navigation predictability.

\begin{itemize}
    \item \textbf{Session-Level User Identification:} given a short browsing session $S$, the goal is to infer the identity of the generating user $u$. This task measures the extent to which persistent behavioral signatures are expressed in short activity fragments. High performance indicates that user-specific preferences are strongly encoded even in limited observations.

    \item \textbf{Next-Domain Prediction:} given a user $u$ and a recent sequence of visited domains $(d_{t-L+1}, \dots, d_t)$, the goal is to predict the next domain $d_{t+1}$. This task captures short-term navigation dynamics and measures how predictable behavior is from recent context combined with long-term interaction history.
\end{itemize}

\subsection{Modeling Approaches}
\label{sec:models}
The modeling instantiations are used to operationalize the tasks introduced above. We distinguish between models designed to capture session-level identity signals and models designed to capture navigation dynamics over user--domain interactions.

\subsubsection{Session-Based Identity Models}
Each session is encoded as a fixed-dimensional feature vector combining behavioral statistics, domain-level signals, and demographic attributes, from which the models estimate a posterior $p(u \mid S)$ over users. We evaluate four families of increasing expressive capacity:

\begin{itemize}
    \item \textbf{Linear Support Vector Machine (SVM):} a linear classifier used to assess whether identity-related signals are linearly separable in the session feature space, serving as a baseline for the inherent structure of identity information.

    \item \textbf{Random Forest (RF):} a non-linear ensemble model that captures interactions between behavioral features and domain-level signals while remaining robust to sparsity and heterogeneous feature distributions.

    \item \textbf{Multilayer Perceptron (MLP):} a neural network that learns complex non-linear combinations of behavioral and domain features, capturing higher-order dependencies that may encode user-specific behavioral signatures.

    \item \textbf{Autoencoder-Based Model (AE):} a representation-learning approach in which session features are first compressed into a latent space via reconstruction, after which a classification head is trained on the learned representation to evaluate whether identity information is preserved under dimensionality reduction.
\end{itemize}

\subsubsection{Structure-Dominant Hybrid Behavioral Model (SD-HBM)}

We propose the Structure-Dominant Hybrid Behavioral Model (SD-HBM) for next-domain
prediction, which models browsing as a structured interaction process over a heterogeneous
bipartite graph $G = (\mathcal{U} \cup \mathcal{D}, E)$, where edges represent observed user–domain interactions. To
account for repeated visits, edge weights use a log-scaled interaction frequency
$w(u,d) = \log(1 + \mathrm{count}(u,d))$, which limits the influence of highly repetitive interactions
while preserving relative preference strength.

\paragraph{Long-Term Structural Representation}
To encode persistent behavioral preferences, SD-HBM employs a two-layer GraphSAGE encoder operating on the bipartite graph $G$. The encoder performs iterative neighborhood aggregation, allowing each user representation to be computed from both directly connected domains and higher-order connectivity patterns reached through shared domains. This yields structural embeddings that capture long-term interests and stable browsing tendencies, and that place users who visit overlapping sets of domains close together in the embedding space.

\paragraph{Short-Term Sequential Modeling}
Short-term navigation behavior is modeled using a GRU-based sequence encoder applied to the most recent $L$ visited domains. The GRU captures temporal dependencies and transition patterns within user sessions, summarizing the recent trajectory into a single state. To enhance temporal sensitivity, embeddings representing hour-of-day and day-of-week are incorporated into the sequential representation. In addition, the embedding of the last visited domain is explicitly concatenated to preserve immediate transition signals, which are often strongly predictive in browsing data.

\paragraph{LLM-derived semantic feature representation}
To complement structural and sequential signals, SD-HBM incorporates an LLM-based feature extraction module. The learned graph-based user embedding is provided to the LLM under a constrained structured prompt, and the model returns a compact set of semantic descriptors summarizing user behavior: dominant and secondary browsing categories, behavioral diversity, routine strength, and category-level affinities such as social, news, productivity and media. These descriptors provide a higher-level semantic abstraction of the learned interaction representation.

\paragraph{Hybrid Behavioral Fusion}
Finally, all representations are integrated into a single latent vector. The GraphSAGE user embedding, GRU-based sequential embedding, temporal encodings, last-domain embedding, and LLM-derived semantic features are concatenated and passed through a feed-forward prediction head that scores candidate domains. The model is trained in two stages, with the graph encoder pretrained and subsequently frozen, and the sequence head trained for next-domain prediction.

\section{Experimental Setup}
\label{sec:experiments}
\subsection{Dataset Overview}
We conduct our experiments using a large anonymous web browsing dataset that has been made available through Zenodo\footnote{\url{https://zenodo.org/records/4757574}}. The dataset consists of time-stamped browsing sessions that have been captured from a group of users within about a month. The data includes information about the identifier of the users, visited domain names, time stamps, and the active duration of browsing sessions.

Additionally, demographic features of the users and a domain-to-category map are provided. Auxiliary features like demographic details and the domain category map are exclusively used for generating features for the session-based identification task and are never used to build future information in the prediction task. After filtering out incomplete records, we obtain a final dataset of more than nine million browsing events from over two thousand users. The dataset is released publicly in anonymized form: users are represented by opaque panel identifiers, and no URLs, page content, or directly identifying attributes are included. Our analysis operates exclusively on domain-level records.

\subsection{Preprocessing and Session Construction}
All experiments are based on a common preprocessing pipeline applied to raw browsing event logs. Events missing a user identifier, domain, or timestamp are removed, and the remaining records are sorted chronologically per user. Sessions are then constructed using a time-gap heuristic: a new session is initiated whenever the interval between consecutive events for the same user exceeds a threshold $\Delta$. This procedure yields temporally coherent sequences that reflect both short-term navigation patterns and longer-term browsing behavior.

We adopt different sessionization thresholds for the two tasks to reflect their distinct temporal requirements. For session-level user identification, we evaluate short inactivity gaps ranging from 1 second to 30 minutes and select a 2-second threshold, which preserves fine-grained behavioral fragments while minimizing noise from unrelated actions. At this resolution, a session corresponds to a tightly grouped burst of requests, which may
include resources loaded alongside a single deliberately visited page. For next-domain prediction, we test thresholds between 2 seconds and 120 seconds and select 30 seconds, which provides a balance between sequence coherence and sufficient length for predictive modeling.

Threshold selection was carried out during pipeline development as an exploratory design choice based on preliminary runs; sessionization parameters are fixed before model training and are not tuned against the reported test results.

\subsection{Task Construction}

\paragraph{Task 1: Session-Level User Identification.}
Each session is encoded into a fixed-dimensional feature vector. Activity features summarize the volume and intensity of the session, including the number of visits, the number of distinct domains visited, and the total and average active time. Temporal features record when the session occurred, using hour of day and day of week together with cyclical encodings. Repetition features describe how concentrated a session is on a small number of domains. Visited domains are represented in two complementary ways: explicit visit counts for the most frequently visited domains in the corpus, and a hashed representation that maps the remaining long tail into a fixed number of bins, preserving domain-level information without an unbounded feature space. Category features aggregate visits through the domain-to-category map, including per-category counts and shares, the number of distinct categories, and the entropy of the category distribution. A small number of cross-session features relate each session to the immediately preceding session of the same user, namely the elapsed time between them and the overlap of their domain and category sets. Finally, demographic attributes are encoded as indicator variables. The dataset is thus composed of sample sessions coupled with user identities. The prediction task is modeled as a multi-class classification over users. To limit class imbalance across users with very different activity levels, we retain the first 50 sessions per user and discard users with fewer than two sessions. This yields 2,140 users and 103,314 sessions for this task, and accounts for the difference in user counts between the two tasks. Table~\ref{tab:task1_dataset} presents the main properties of Task~1 dataset.

\begin{table}[htbp]
\begin{center}
\caption{Dataset characteristics for Task 1 (Session-Level User Identification).}
\label{tab:task1_dataset}
\begin{tabular}{l c}
\hline
\textbf{Characteristic} & \textbf{Value} \\
\hline
Time span & October 2018 \\
Number of users (classes) & 2{,}140 \\
Number of sessions & 103{,}314 \\
Number of raw browsing events & $\sim$9 million \\
Session definition & Time-gap rule (2 seconds) \\
Train/Test split & 80\% / 20\% (stratified by user) \\
\hline
\end{tabular}
\end{center}
\end{table}

\paragraph{Task 2:  next-domain prediction.}
In order to predict the next domain in the navigation sequence, supervised samples are generated through a sliding window over sessions. Based on previously visited $L$ domains, the model makes predictions over the next domain. In order to make the prediction easier by limiting the size of the output space, only the top $K$ frequently visited domains are considered in the training data.  Table~\ref{tab:task2_dataset} reports the main statistics of the Task~2 dataset. Domains outside this vocabulary are mapped to a single UNK class, which groups the long tail of rarely visited sites. UNK is treated as an ordinary prediction target rather than being excluded from evaluation: samples whose true next domain falls outside the vocabulary remain in the test set and are counted in the accuracy denominator. Supervised samples are generated only from sessions containing more than $L$ events, so the reported predictability characterizes sustained browsing episodes rather than the full session population.

\begin{table}[htbp]
\begin{center}
\caption{Dataset characteristics for Task 2 ( next-domain prediction).}
\label{tab:task2_dataset}
\begin{tabular}{l c}
\hline
\textbf{Characteristic} & \textbf{Value} \\
\hline
Number of users & 2{,}148 \\
Number of raw browsing events & 9{,}151{,}243 \\
Number of sessions (gap = 30s) & 2{,}215{,}002 \\
Sequence length $L$ & 10 \\
Number of supervised samples & 356{,}707 \\
Target vocabulary size & Top-10{,}000 + UNK \\
UNK fraction & 0.78\% \\
Split strategy & Chronological per user \\
\hline
\end{tabular}
\end{center}
\end{table}

\subsection{Implementation Details}
We conduct empirical hyperparameter tuning over architectural and optimization configurations. For session-level user identification, a validation set is held out from the training partition and used for model selection and early stopping. For next-domain prediction, configurations are selected on the held-out evaluation split under a fixed training budget, as described below.

\subsubsection{Task 1: Session-Level User Identification}
SVM and RF use standard library implementations, for classical machine learning baselines. The best-performing model is a
residual MLP with seven fully connected layers. Layers 1 to 5 each contain 1024 hidden neurons, followed by a sixth layer with 512 neurons. The final layer maps to the user classes. Each hidden layer is followed by ReLU activation, Batch Normalization (BatchNorm1d), and Dropout with a rate of 0.2 to reduce overfitting. The model is trained using the Cross-Entropy loss with label smoothing set to 0.05. Optimization is performed using Adam with a learning rate of 0.001, batch size of 1024, and a maximum of 200 training epochs. Early stopping is applied with a patience of 8 epochs based on validation Top-1 accuracy.

\subsubsection{Task 2:  next-domain prediction (SD-HBM)}
Structure-Dominant Hybrid Behavioral Model (SD-HBM) extends the Graph-Sequence Model (GSM) with LLM-based semantic feature augmentation.

A two-layer GraphSAGE encoder is first applied over the constructed interaction graph $G$ to learn 128-dimensional user and domain embeddings. The encoder is trained using a pairwise ranking objective over observed edges with negative sampling (five negatives per positive interaction). Optimization is performed using AdamW with a learning rate of 0.001 for 5 epochs, a fixed budget selected during development; the encoder is subsequently frozen and used as a static representation.

Then, short-term navigation behavior is modeled using a GRU encoder over the last $L = 10$ visited domains, each represented by a 128-dimensional embedding. The GRU has one layer with hidden size 128, and its final hidden state is used as the sequence representation. We additionally include contextual features consisting of the 128-dimensional user embedding, the last-domain embedding, and temporal embeddings for hour-of-day (16D) and day-of-week (8D).

All components are concatenated into a 408-dimensional representation. When LLM-derived semantic features are included, the representation dimension increases to 418. The fused vector is projected into a 128-dimensional space using a fully connected layer with GELU and 0.2 dropout, followed by L2 normalization. The model is trained using an InfoNCE / sampled softmax objective optimized with AdamW (learning rate 0.002, batch size 4096) and gradient clipping with norm 1.0.

For LLM-based semantic feature generation, we augment the model with semantic features generated by GPT-5.4-mini, using the provider's default decoding settings. Generation is constrained by a strict JSON schema that fixes the set of output keys and restricts numeric fields to $[0,1]$, which bounds the output space; we did not measure variability across repeated calls for the same input.

To contextualize the performance of the proposed SD-HBM, we evaluate three complementary baseline approaches under the same experimental setting. The Graph-Based Markov Model (GBM) defines a probabilistic random walk over the user--domain interaction graph, where transition probabilities are estimated from normalized visit counts. Predictions are obtained by a random walk with restart, where the restart distribution is concentrated on the current user and on the most recently visited domains, with greater weight assigned to more recent visits. The Graph-Sequence Model (GSM) combines GraphSAGE-based user and domain embeddings with a GRU-based sequence encoder to jointly model long-term interaction structure and short-term navigation dynamics without semantic augmentation. In addition, we consider a semantic augmentation baseline, referred to as the GSM + LLM Reranking, where the GSM first produces a ranked list of candidate next domains, and a LLM is then used to rerank these candidates based on the recent browsing sequence, user context, and optional domain category information by estimating their semantic plausibility. All next-domain models are trained on chronologically ordered splits, and the interaction graph is constructed from training sessions only. Model selection for this task is performed on the held-out evaluation split, so the reported figures correspond to the best epoch under a fixed training budget rather than to an independently validated configuration. The same protocol is applied to every model in this task, which keeps the comparison internally consistent. Standard regularization techniques, including early stopping, dropout, and gradient clipping, are applied consistently across neural models to ensure stable training and fair comparison. In terms of computational cost, the classical identification baselines train on CPU within minutes, whereas the neural and graph-based models require GPU acceleration; graph encoder pretraining is the most expensive stage, and inference in all cases is a single forward pass.

\subsection{Evaluation Metrics}
Evaluation of all models across both tasks is performed based on Top-1 accuracy to ensure comparability of inference performance under a unified evaluation criterion. It measures the ratio of correctly predicted samples to the total number of evaluated samples. The accuracy is defined for a set of predicted labels $\hat{y}_i$ and actual labels $y_i$ by:

\begin{equation}
\text{Top-1 accuracy} = \frac{1}{N} \sum_{i=1}^{N} \mathbb{I}[\hat{y}_i = y_i].
\end{equation}
where $N$ is the total number of samples.

Top-1 accuracy is a strict criterion in both settings: the label space contains 2,140 users for identification and 10,001 classes for next-domain prediction, so a uniform random predictor would achieve well below one percent in either task. We report Top-1 only in order to keep a single comparable criterion across model families that produce scores in different ways. Richer ranking metrics, such as Recall@$k$ or mean reciprocal rank, would give a fuller picture of how candidates are ordered below the top position, and we leave their systematic reporting to future work.

\subsection{Evaluation Protocol}
Our experimental design aims to avoid temporal and structural leakage during training and evaluation. The following constraints are enforced:

\begin{itemize}
    \item For next-domain prediction, sessions are ordered chronologically per user and split so that earlier sessions are used for training and later sessions are held out for evaluation. For session-level user identification, sessions are partitioned using a stratified random split over users (Table I), which preserves the class distribution across all identities. Because this split is not chronological, sessions recorded close together in time may fall on opposite sides of the partition, and the reported identification accuracy should therefore be interpreted as an upper bound relative to a strictly temporal protocol.

    \item Graph construction from training data only: Graphs used for navigation task are generated solely using training sessions; there are no test-time interactions in these graphs.

    \item No future information in features:  Session-level features are computed from events within the session and from preceding sessions only; no feature depends on activity that occurs later in the same user's trace.

    \item Consistent evaluation protocol: Models are all evaluated using the same test splits across all statistical, neural, graph-based, and LLM-based approaches.
    
 \end{itemize}
\section{Experimental Results}
\label{sec:results}
This section presents empirical findings for both tasks. Results are interpreted based on their relevance to understanding the underlying nature of the web browsing activity. Specifically, we analyze (i) whether short browsing sessions are identifiable, (ii) how predictable navigation behavior is, and (iii) whether semantic augmentation via LLMs provides additional behavioral signal.

\subsection{Are Browsing Sessions Identifiable?}

First, we look at whether short browsing sessions are enough to enable identifying the generating users. Table~\ref{tab:task1} summarizes the performance of all models in terms of user identification at the session level, depending on model expressiveness. There is a noticeable trend that can be observed here. Specifically, linear models have poor results (30.5\% accuracy), which implies that the identity signals cannot be separated using linear algorithms. Nonetheless, the performance grows with increasing complexity, with tree models obtaining impressive results (84.8\%) and neural models reaching the best performance, with MLP getting 89.8\% of Top-1 accuracy.

This indicates that the behavioral patterns are already sufficiently distinct even for short periods of web navigation. Most importantly, there is no single feature that holds identity-related signals – the information about users is extracted from complicated interactions between behavior, domains, and demographics.

\subsection{How Predictable is Navigation Behavior?}
We next analyze the predictability of short-term navigation behavior. Table~\ref{tab:task2_graph} reports results for graph-based, sequential, and LLM-enhanced models. The GBM model achieves 71.44\% accuracy,  showing that transition statistics over the interaction graph already capture pronounced repetition in user navigation. Because the walk restarts predominantly from recently visited domains, this figure reflects a combination of persistent user--domain structure and short-term recency rather than long-term structure alone.

Building on this, the GSM model improves performance to 79.38\%, demonstrating the benefit of combining long-term structural embeddings with short-term sequential modeling.  

Further gains are observed when incorporating LLM-based components. The GSM + LLM Reranking model achieves 79.84\%, showing that semantic reordering of candidate domains can refine predictions produced by the structural model. Finally, the proposed SD-HBM model achieves the best performance with 80.15\%, indicating that integrating semantic features directly into the representation space is more effective than post-hoc reranking.

\subsection{Do LLMs Add Behavioral Information?}
Across all configurations, LLM-based components provide consistent but small improvements
over the strongest non-LLM baseline (GSM, 79.38\%): reranking yields 79.84\%, and feature
augmentation reaches 80.15\% (SD-HBM). Semantic features therefore contribute complementary
information, but the dominant predictive signal remains encoded in the interaction graph and
recent behavioral sequences, and integrating semantic features into the representation is more
effective than post-hoc reranking.

\begin{table}[H]
\caption{Session-level user identification performance (Task 1).}
\begin{center}
\label{tab:task1}
\begin{tabularx}{\columnwidth}{X c}
\toprule
\textbf{Model} & \textbf{Top-1 Accuracy} \\
\midrule
MLP & \textbf{0.8978} \\
Random Forest & 0.8480 \\
Autoencoder (AE) & 0.8068 \\
SVM & 0.3050 \\
\bottomrule
\end{tabularx}
\end{center}

\end{table}

\begin{table}[H]
\begin{center}
\caption{ next-domain prediction performance (Task 2).}
\label{tab:task2_graph}
\begin{tabular}{lc}
\toprule
\textbf{Model} & \textbf{Top-1 Accuracy} \\
\midrule

Graph-Based Markov Model (GBM) & 0.7144 \\
Graph-Sequence Model (GSM) & 0.7938 \\
GSM + LLM Reranking & 0.7980 \\
SD-HBM  & \textbf{0.8015} \\
\bottomrule
\end{tabular}
\end{center}
\end{table}

\section{Discussion}

\label{sec:discussion}

Our results suggest that user identifiability and navigation predictability are driven by a shared underlying interaction structure rather than fundamentally distinct behavioral processes. Across both tasks, repeated user--domain interactions emerge as the dominant source of signal, indicating that behavioral traces encode stable identity-related and short-term predictive information within the same representational space.

From a modeling perspective, the strong performance of graph-based and sequence-based approaches indicates that most of the exploitable structure in browsing behavior is already captured through interaction history and temporal ordering. Sequential modeling provides additional gains by refining local dynamics, but does not fundamentally alter the predictive capacity provided by structural representations.

The limited impact of LLM-derived semantic feature augmentation further suggests that high-level semantic interpretations of behavior add only marginal information beyond what is already encoded in interaction patterns. In particular, semantic features appear to refine rather than transform the underlying representation, indicating that structural interaction patterns contribute substantially more predictive information than the evaluated LLM-derived semantic features.

From a privacy perspective, these results indicate that domain-level browsing traces carry a persistent identity signal: a single short session is often sufficient to recover the generating user among more than two thousand candidates, without access to URLs, page content, or explicit identifiers. The panel data used here is public and anonymized, yet anonymization at the identifier level evidently does not remove behavioral linkability.
Plausible mitigations therefore operate on the representation rather than the identifier: aggregating visits to the category level, coarsening timestamps, or suppressing rare domains would each reduce the distinctiveness that our features exploit. We do not evaluate such defenses here, and measuring identification accuracy under adversarial or privacy-preserving transformations remains an important direction for future work.

\section{Limitations}
\label{sec:limitations}
Despite these findings, several limitations should be acknowledged. First, the analysis is based on a single dataset collected over a limited time period, which restricts the ability to capture long-term behavioral evolution, seasonal effects, or cross-period generalization.

Second, the study operates at the domain level, without incorporating page-level content, search queries, or intra-site navigation structure. While this abstraction improves scalability and reduces noise, it necessarily omits finer-grained behavioral signals that may further improve both identity and intent modeling.

Third, the two tasks use different evaluation protocols. Next-domain prediction is evaluated
under a chronological split, whereas user identification uses a stratified random partition
over users. The identification result is therefore not directly comparable to a strictly
temporal evaluation and should be read as an upper bound.

Fourth, session construction at a 2-second threshold groups requests occurring in rapid
succession. Such groups may reflect resources co-loaded with a single deliberate visit rather
than a sequence of navigation decisions, so identification at this resolution partly reflects
page-loading footprints in addition to user choices.

Fifth, supervised samples for next-domain prediction are drawn only from sessions longer than
$L$ events, so the reported predictability characterizes sustained browsing episodes rather
than browsing activity as a whole. The autoencoder is likewise evaluated at a single latent
size, so we do not characterize how identification accuracy degrades under stronger
compression.

Sixth, all reported results correspond to single training runs. We do not report variance across random seeds or statistical tests, and the sub-one-point differences separating the strongest next-domain models are within the range plausibly attributable to run-to-run variation; their ordering should therefore be treated as indicative rather than conclusive. Relatedly, we do not include trivial reference baselines such as repeating the last visited domain or predicting each user's most frequently visited domain, which would help establish the margin achieved over simple heuristics.

Seventh, the identification features include demographic attributes, which are constant per user. The reported accuracy therefore reflects a combination of behavioral and demographic signal, and isolating the purely behavioral component is left to future work.

Finally, modeling choices such as sessionization thresholds, Top-K vocabulary restriction,
and session filtering may still influence the observed performance.

%Finally, the framework assumes relatively stable user identities over time. In practice, user behavior may vary across devices, contexts, or behavioral modes, which is not explicitly modeled in this study.

\section{Conclusion}
\label{sec:conclusion}

In this work, we investigated user identifiability and navigation predictability as two complementary inference tasks derived from web browsing traces. We used a common experimental framework, with task-specific preprocessing, to study how different modeling families exploit interaction patterns in user–domain behavior. Our findings show that both identity inference and next-action prediction are largely governed by repeated interaction structures between users and domains. Structural and sequential models capture most of the informative signal, while semantic augmentation using LLMs provides only limited additional benefit. Overall, the results indicate that browsing behavior is primarily shaped by interaction history and repetition patterns, with semantic interpretations playing a secondary role. This supports the view that both identity-related and predictive signals in web browsing appear to be strongly influenced by similar interaction regularities.

\vspace{12pt}
\bibliographystyle{IEEEtran}
\bibliography{ref}

@inproceedings{Kulshrestha2021ICWSM,
  author    = {Juhi Kulshrestha and Marcos Oliveira and Orkut Kara{\c{c}}alik and Denis Bonnay and Claudia Wagner},
  title     = {Web Routineness and Limits of Predictability: Investigating Demographic and Behavioral Differences Using Web Tracking Data},
  booktitle = {Proceedings of the International AAAI Conference on Web and Social Media (ICWSM)},
  year      = {2021},
  pages     = {327--338},
  doi       = {10.1609/icwsm.v15i1.18064}
}

@article{Wang2021CSURSessionSurvey,
  author  = {Shoujin Wang and Longbing Cao and Yan Wang and Quan Z. Sheng and Mehmet A. Orgun and Defu Lian},
  title   = {A Survey on Session-based Recommender Systems},
  journal = {ACM Computing Surveys},
  year    = {2021},
  volume  = {54},
  number  = {7},
  pages   = {1--38},
  articleno = {154},
  doi     = {10.1145/3465401}
}

@article{LatifiMauroJannach2021InfSci,
  author  = {Sara Latifi and Noemi Mauro and Dietmar Jannach},
  title   = {Session-aware Recommendation: A Surprising Quest for the State-of-the-art},
  journal = {Information Sciences},
  year    = {2021},
  volume  = {573},
  pages   = {291--315},
  doi     = {10.1016/j.ins.2021.05.048}
}

@inproceedings{CuiEtAl2021MotifSIGIR,
  author    = {Zeyu Cui and Yinjiang Cai and Shu Wu and Xibo Ma and Liang Wang},
  title     = {Motif-aware Sequential Recommendation},
  booktitle = {Proceedings of the 44th International ACM SIGIR Conference on Research and Development in Information Retrieval (SIGIR)},
  year      = {2021},
  pages     = {1738--1742},
  doi       = {10.1145/3404835.3463115}
}

@article{LatifiJannachFerraro2022InfSci,
  author  = {Sara Latifi and Dietmar Jannach and Andr{\'e}s Ferraro},
  title   = {Sequential Recommendation: A Study on Transformers, Nearest Neighbors and Sampled Metrics},
  journal = {Information Sciences},
  year    = {2022},
  volume  = {609},
  pages   = {660--678},
  doi     = {10.1016/j.ins.2022.07.079}
}

@article{Boka2024SeqRecSurvey,
  author  = {T. F. Boka and R. B. Neupane and Z. Niu},
  title   = {A Survey of Sequential Recommendation Systems: Techniques, Evaluation, and Future Directions},
  journal = {Information Systems},
  year    = {2024},
  volume  = {125},
  pages   = {102427},
  doi     = {10.1016/j.is.2024.102427}
}

@article{Pan2026SeqRecSurvey,
  author  = {Li-Wei Pan and Wei-Ke Pan and Mei-Yan Wei and Hong-Zhi Yin and Zhong Ming},
  title   = {A Survey on Sequential Recommendation},
  journal = {Frontiers of Computer Science},
  year    = {2026},
  volume  = {20},
  number  = {3},
  pages   = {2003606},
  doi     = {10.1007/s11704-025-41329-w}
}

@article{BlackEtAl2024TRACE,
  author  = {William Black and Alexander Manlove and Jack Pennington and Andrea Marchini and Ercument Ilhan and Vilda Markeviciute},
  title   = {TRACE: Transformer-based User Representations from Attributed Clickstream Event Sequences},
  journal = {CoRR},
  year    = {2024},
  volume  = {abs/2409.12972},
  url     = {https://arxiv.org/abs/2409.12972}
}

@inproceedings{VernaEtAl2024CoNEXT,
  author    = {Alberto Verna and Nikhil Jha and Martino Trevisan and Marco Mellia},
  title     = {A First View of Topics {API} Usage in the Wild},
  booktitle = {Proceedings of the ACM International Conference on Emerging Networking Experiments and Technologies (CoNEXT)},
  year      = {2024},
  doi       = {10.1145/3680121.3697810}
}

@article{BeuginMcDaniel2024TopicsAssessment,
  author  = {Yohan Beugin and Patrick McDaniel},
  title   = {A Public and Reproducible Assessment of the Topics {API} on Real Data},
  journal = {CoRR},
  year    = {2024},
  volume  = {abs/2403.19577},
  url     = {https://arxiv.org/abs/2403.19577}
}

@article{SchnauberStockmann2023Predictability,
  title={Routines and the Predictability of Day-to-Day Web Use},
  author={Schnauber-Stockmann, Anna and others},
  journal={Media Psychology},
  year={2023}
}

@article{Paulino2024WebTraceSense,
  title={Leveraging WebTraceSense for User Interaction Log Analysis and Visualization},
  author={Paulino, Diogo and others},
  journal={ACM Transactions on Interactive Intelligent Systems},
  year={2024}
}

@article{Oliveira2025BehavioralFingerprinting,
  title={Browsing Behavior Exposes Identities on the Web},
  author={Oliveira, M. and others},
  journal={Scientific Reports},
  year={2025}
}

@article{Song2025WebsiteFingerprinting,
  title={Redefining Website Fingerprinting Attacks With Multiagent LLMs},
  author={Song, Chuxu and others},
  journal={arXiv preprint},
  year={2025}
}
\end{document}